%% file: template.tex
\documentclass{article}

\usepackage{arxiv}

\usepackage[utf8]{inputenc} 
\usepackage[T1]{fontenc}    
\usepackage{hyperref}       
\usepackage{url}            
\usepackage{booktabs}       
\usepackage{amsfonts}       
\usepackage{nicefrac}       
\usepackage{microtype}      
\usepackage{lipsum}		
\usepackage{graphicx}
\usepackage{natbib}
\usepackage{doi}
\usepackage{subcaption}

\title{Fairness and/or Privacy on Social Graphs}

\date{} 					

\author{ \hspace{1mm}Bartlomiej Surma, Michael Backes, Yang Zhang\\
	CISPA Helmholtz Center for Information Security\\
	\texttt{\{bartlomiej.surma, backes, yang.zhang\}@cispa.saarland} \\
}

\renewcommand{\headeright}{Technical Report}
\renewcommand{\undertitle}{Technical Report}
\renewcommand{\shorttitle}{Fairness and/or Privacy on Social Graphs}

\hypersetup{
pdftitle={Fairness and/or Privacy on Social Graphs},
pdfsubject={},
pdfauthor={Bartlomiej Surma},
pdfkeywords={GCN, GNN, privacy, fairness},
}

\begin{document}
\maketitle

\begin{abstract}
Graph Neural Networks (GNNs) have shown remarkable success in various graph-based learning tasks. However, recent studies have raised concerns about fairness and privacy issues in GNNs, highlighting the potential for biased or discriminatory outcomes and the vulnerability of sensitive information. This paper presents a comprehensive investigation of fairness and privacy in GNNs, exploring the impact of various fairness-preserving measures on model performance. We conduct experiments\footnote{Source code for all experiments available at \url{https://github.com/EnderGed/FairGCN}} across diverse datasets and evaluate the effectiveness of different fairness interventions. Our analysis considers the trade-offs between fairness, privacy, and accuracy, providing insights into the challenges and opportunities in achieving both fair and private graph learning. The results highlight the importance of carefully selecting and combining fairness-preserving measures based on the specific characteristics of the data and the desired fairness objectives.  This study contributes to a deeper understanding of the complex interplay between fairness, privacy, and accuracy in GNNs, paving the way for the development of more robust and ethical graph learning models.
\end{abstract}

\input{1_introduction.tex}
\input{2_relatedwork.tex}
\input{3_dataset.tex}
\input{4_methodology.tex}
\input{5_findings}
\input{6_conclusion}

\newpage

\bibliographystyle{unsrtnat}
\bibliography{references}  

\end{document}

%% file: 1_introduction.tex
\section{Introduction}

Graph Neural Networks (GNNs) have emerged as a powerful tool for learning representations of graph-structured data, achieving remarkable success in various domains, including social network analysis, recommendation systems, and drug discovery \cite{wu2020comprehensive,zhou2020graph}. However, as GNNs become increasingly prevalent in real-world applications, concerns regarding fairness and privacy have come to the forefront. GNNs, like other machine learning models, can inherit and amplify biases present in the training data, leading to discriminatory outcomes or unfair treatment of certain groups \cite{barocas2016big,mehrabi2021survey}. Moreover, the rich relational information captured by GNNs can potentially expose sensitive or private information about individuals or entities represented in the graph.

Addressing fairness and privacy concerns in GNNs is crucial for ensuring responsible and ethical use of these powerful models. This paper presents a comprehensive investigation of fairness and privacy in GNNs, exploring the impact of various fairness-preserving measures on model performance. We conduct experiments across diverse datasets and evaluate the effectiveness of different fairness interventions, considering their effects on accuracy, privacy, and fairness metrics. Our analysis provides valuable insights into the trade-offs between these competing objectives, guiding the development of more robust and ethical graph learning models.

%% file: 2_relatedwork.tex
Research on fairness in machine learning has gained significant attention in recent years, with numerous studies focusing on mitigating bias and ensuring equitable outcomes \cite{barocas2016big,mehrabi2021survey}. Several approaches have been proposed to address fairness concerns, including pre-processing techniques that modify the training data \cite{kamiran2012data}, in-processing methods that incorporate fairness constraints during model training \cite{zemel2013learning}, and post-processing techniques that adjust model outputs to achieve fairness \cite{hardt2016equality}.

In the context of GNNs, fairness research is still in its early stages but has seen growing interest. Existing work has explored various fairness interventions for GNNs, such as adversarial debiasing \cite{zhang2018mitigating}, fair regularization techniques \cite{bose2019compositional}, and modifying the graph structure to promote fairness \cite{rahman2019fairwalk}. However, there is a lack of comprehensive studies that systematically evaluate the impact of different fairness-preserving measures on GNN performance, considering the interplay between fairness, privacy, and accuracy.

Privacy in Graph Neural Networks (GNNs) is a growing concern due to the sensitive nature of graph data, which often contains personal information and relationships \cite{zhang2024survey}. GNNs, while powerful in learning representations, can inadvertently leak or expose this sensitive information through their predictions or learned embeddings \cite{yuan2024can}. This can lead to privacy violations, such as membership inference attacks, where an adversary can determine if a specific node was part of the training data, or attribute inference attacks, where sensitive attributes of individuals are revealed \cite{olatunji2023does}. These privacy risks underscore the critical need for developing and integrating privacy-preserving techniques into GNN models to safeguard sensitive information and ensure responsible use of graph data.

This paper builds upon the existing literature by conducting a comprehensive analysis of fairness and privacy in GNNs. We evaluate a range of fairness interventions and privacy metrics, providing valuable insights into the trade-offs and challenges in achieving both fair and private graph learning. Our study contributes to the growing body of knowledge on responsible and ethical AI, particularly in the context of graph-structured data.

%% file: 3_dataset.tex
\section{Experimental Setup}
\subsection{Dataset}
The study utilizes 2 social network datasets with personal attributes. Table~\ref{tab:dataset} shows the basic statistics of the datasets. For each graph we're splitting

\paragraph{NBA}
Kaggle dataset\footnote{https://www.kaggle.com/datasets/noahgift/social-power-nba}, encompassing approximately 400 professional basketball players. The dataset provides comprehensive performance statistics from the 2016-2017 NBA season, alongside demographic information including nationality, age, and salary. To construct a player network, Twitter follower relationships were collected using the official Twitter API. This network represents connections between players based on their Twitter interactions.The primary classification task involves predicting whether a player's salary exceeds the median salary within the dataset.

We treat age (one of 4 almost equal age groups) as a privacy sensitive feature, country (from US / from abroad) as a fairness sensitive attribute, and salary (above / below median) as the classification task.

\paragraph{Pokec}
Introduced by \citep{takac2012data}, representing the largest online social network in Slovakia. Pokec has maintained its popularity for more than a decade, connecting more than 1.6 million users. The dataset provides anonymized user profiles, including demographic information such as gender, age, hobbies, interests, and education, all presented in the Slovak language. Additionally, it contains the directed friendship network representing user connections.
For the purpose of this research, we performed post-processing to create subgraphs focused on specific geographic regions. Following methodology by /citep{dai2021say}, we subsampled two subgraphs pokec\_z and pokec\_n

We treat gender as a privacy sensitive attribute, location (from one of the 2 major regions) as a fairness sensitive attribute,  and occupation (one of 4 or 5 most popular jobs) as the classification task.

\begin{table}[!t]
\centering
\begin{tabular}{r|l|l|l}
\toprule
Dataset & nba & pokec\_n & pokec\_z \\
\midrule
\# nodes & 400 & 66.6k & 67.8k \\
\# edges & 10.6k & 517k & 618k \\
\# node attributes & 95 & 265 & 276 \\
\% nodes labeled & 77.5 & 13.2 & 15.1 \\
\bottomrule
\end{tabular}
\caption{Dataset statistics \label{tab:dataset}}
\end{table}

\subsection{Fairness definitions}
Fairness in machine learning is a multifaceted concept with various definitions, each addressing different aspects of discriminatory outcomes. \citet{alikhademi2022review, wang2022brief, kusner2019making}  In this paper, we focus on \textit{Statistical Parity} and \textit{Equality of Opportunity}.
\paragraph{Statistical Parity} also known as demographic parity \citep{dwork2012fairness} , is a fairness criterion used in machine learning to assess whether a model's outcomes are equally distributed across different demographic groups.

In essence, it aims to ensure that the probability of a positive outcome is the same for all protected groups, regardless of their sensitive attributes (such as race, gender, or age).

If $\hat{Y}$ represents the predicted label and $A$ represents the protected attribute, then Statistical Parity requires:
$$P(\hat{Y} = 1 | A = a) = P(\hat{Y} = 1 | A = a') \quad \forall a, a' \in \mathcal{A}$$

\paragraph{Equality of Opportunity} \citep{hardt2016equality} is defined as a fairness criterion that requires a classifier to have equal true positive rates across different protected groups.

If $Y$ represents the true label, $A$ represents the protected attribute, and $\hat{Y}$ represents the predicted label, then Equality of Opportunity requires:
$$P(\hat{Y} = 1 | Y = 1, A = a) = P(\hat{Y} = 1 | Y = 1, A = a') \quad \forall a, a' \in \mathcal{A}$$

\paragraph{Fairness Attribute Leakage}

To assess the extent to which sensitive attributes are encoded within the learned node embeddings, we quantify fairness attribute leakage. This is achieved by training a separate, simple model (e.g., a logistic regression classifier) that aims to predict the sensitive attribute solely from the node embeddings. The accuracy of this predictive model serves as a measure of fairness attribute leakage. Higher accuracy indicates greater leakage, suggesting that the sensitive attribute is readily discernible from the embeddings, potentially leading to privacy violations or discriminatory outcomes.

\subsection{Privacy Definitions}

Protecting privacy in machine learning is crucial to prevent the inadvertent leakage or misuse of sensitive information contained within the training data. To assess the privacy risks associated with the trained models, we consider the following privacy metrics:

\paragraph{Privacy Attribute Leakage}

Analogous to fairness attribute leakage, we quantify privacy attribute leakage by training a separate model to predict privacy-sensitive attributes from the learned node embeddings. The accuracy of this predictive model serves as a measure of privacy leakage, with higher accuracy indicating greater vulnerability to privacy attacks.

\paragraph{Membership Inference Attack}

Membership inference attacks aim to determine whether a specific data point was used in the training data. To evaluate the susceptibility of the models to such attacks, we train two distinct attack models: a simple model using XGBoost and a more complex multi-layered neural network. The accuracy of these attack models in identifying members of the training data will be reported as a measure of the models' vulnerability to membership inference attacks.

%% file: 4_methodology.tex
\section{Methodology}
\subsection{Graph Convolution Networks}
Graph Convolutional Networks (GCNs) are a class of neural networks designed to operate on graph-structured data. Unlike traditional convolutional neural networks (CNNs) that excel at processing grid-like data such as images, GCNs leverage the inherent relationships between nodes in a graph to learn meaningful representations. The core idea behind GCNs is to aggregate information from a node's neighbors, effectively "convolving" the node's features with the features of its connected nodes. This process allows the network to capture both the individual node attributes and the structural information encoded in the graph's connectivity. \textbf{GCN}s were popularized by \citep{kipf2016semi}, which introduced an efficient layer-wise propagation rule for graph convolutions, simplifying earlier spectral-based approaches.

Further advancements in graph neural networks have addressed the limitations of the original GCN framework. 
\paragraph{GraphSAGE} \citep{hamilton2017inductive} addressed the limitation of inductive learning on large graphs. Instead of learning a fixed embedding for each node, GraphSAGE learns aggregator functions that sample and aggregate features from a node's local neighborhood. This inductive capability allows GraphSAGE to generalize to new, unseen nodes, making it suitable for dynamic and evolving graphs.
\paragraph{Graph Attention Networks} (GATs) \citep{velivckovic2017graph} introduced an attention mechanism to the aggregation process, allowing the network to learn the importance of different neighbors. By assigning attention weights to neighboring nodes, GATs can focus on the most relevant information during aggregation, improving the model's ability to capture complex dependencies and handle noisy or irrelevant neighbors. This attention mechanism also allows for a more interpretable model, as the attention weights provide insights into which neighbors contributed most to a node's representation.
\paragraph{Graph Isomorphism Networks} (GINs) \citep{xu2018powerful} focus on maximizing the discriminative power of GNNs. GINs are designed to be as powerful as the Weisfeiler-Lehman (WL) graph isomorphism test, a powerful method for distinguishing graphs. They achieve this by employing an injective aggregation function that uses multi-layer perceptrons (MLPs) to learn how to effectively aggregate neighboring node features. This allows GINs to differentiate between graphs that standard GCNs and other GNN architectures might fail to distinguish, making them particularly well-suited for tasks that require a high level of discriminative power.

Graph Convolutional Networks (GCNs) generate node embeddings by iteratively aggregating and transforming feature information from a node's local neighborhood. This process leverages the graph structure to capture both node attributes and relational dependencies. Specifically, GCN layers propagate information by averaging or summing the features of neighboring nodes, followed by a non-linear activation. The resulting embeddings encode a node's position within the network and its inherent characteristics, effectively summarizing its local graph environment. These learned embeddings serve as powerful representations that can be utilized for various downstream inference tasks. In our work, we leverage these GCN-generated node embeddings to analyze fairness and privacy vulnerabilities. By using the embeddings as input to fairness evaluation metrics and privacy attack models, we aim to demonstrate how biases and sensitive information can be exposed or exploited within the graph-structured data. This approach allows us to assess the robustness of graph learning models against fairness and privacy threats, highlighting the importance of developing robust and ethical graph analysis techniques.

\subsection{Classification task}
\begin{figure}[]
    \centering
    \includegraphics[width=0.7\textwidth]{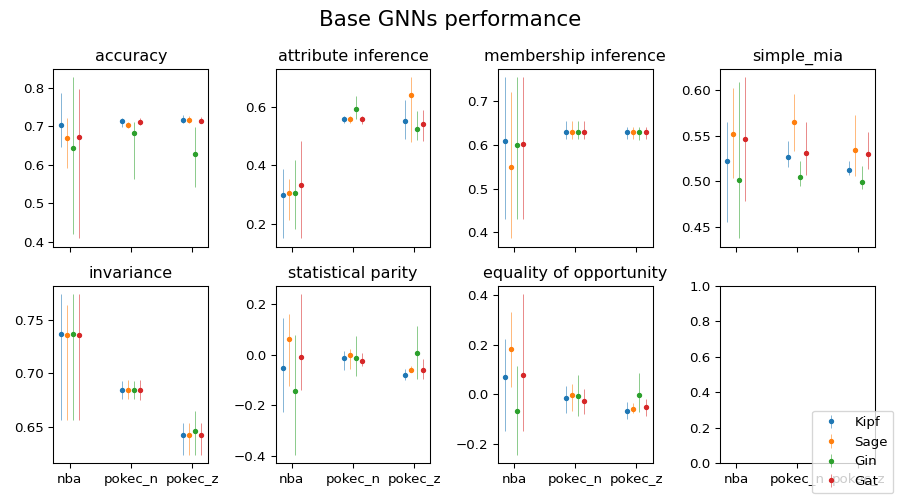}
    \caption{Base model performance on all metrics, averaged with standard deviation for different test / train splits}
    \label{fig:base_performance}
\end{figure}

For each graph, we divide the classification labels (salary or occupation) into training (50\%), validation (25\%) and testing (25\%). 
For each network GCN, GraphSAGE and GAT we train in binary classification task, perform hyper-parameter tuning in the validation set, and report the accuracy and other metrics discussed in this paper in figure~\ref{fig:base_performance}.

\subsection{Fairness countermeasure methods}
\paragraph{Adversarial de-biasing} \citep{dai2021say} proposes a method to mitigate discrimination in Graph Neural Networks (GNNs) when sensitive attribute information is partially or fully unavailable. The approach involves learning a proxy sensitive attribute representation by leveraging the graph structure and available node features. This proxy is then used to regularize the GNN training process, aiming to minimize the correlation between node embeddings and the inferred sensitive attribute. Specifically, the method introduces a fairness regularizer that encourages the GNN to produce node embeddings that are independent of the learned proxy sensitive attribute, thus reducing bias in the resulting representations.
\paragraph{Embedding projection} \citep{palowitch2019monet} modifies the graph embedding training process by incorporating a regularization term into the loss function. This term calculates the inner product between the node embeddings and vectors representing sensitive metadata. The optimization process then minimizes this inner product, effectively reducing the correlation between the learned node embeddings and the sensitive attributes. This minimization is intended to produce embeddings that are less dependent on the sensitive information.
\paragraph{Edge weightening} FairWalk \citep{rahman2019fairwalk} introduced a modification of node2vec \citep{grover2016node2vec}. They attach weights to edges based on the inverse of proportions of sensitive groups, to make it more likely to visit under-represented nodes during random walk part of node2vec. We extend the idea for GCNs, where we use the edge weights during the convolution step of GCNs.
\paragraph{Filtering} \citep{bose2019compositional} introduces a framework for enforcing fairness in graph embeddings by defining compositional fairness constraints. This framework allows for the specification of complex fairness requirements involving multiple sensitive attributes and logical combinations of them. The method achieves fairness by incorporating these constraints into the graph embedding loss function as regularization terms. Specifically, it constructs penalties based on the violations of the defined fairness constraints, encouraging the model to learn embeddings that satisfy the desired fairness properties. This approach provides a flexible way to enforce various fairness notions, including those that consider intersections and combinations of sensitive attributes.
\paragraph{Fair learning} \citep{zhang2018mitigating} proposes a method for debiasing representations using adversarial training.  The method involves training two neural networks: a predictor and an adversary. The predictor learns to perform the main task (e.g., classification) while the adversary tries to predict a protected attribute (e.g., race, gender) from the predictor's output. By training these networks against each other, the predictor learns to achieve its task while making it difficult for the adversary to infer the protected attribute. This adversarial process encourages the predictor to learn representations that are disentangled from the protected attribute, thus mitigating bias.

The debiasing process is contingent upon the specific fairness definition being optimized. Therefore, we introduce distinct nomenclature to differentiate between models trained with different fairness objectives. Specifically, we denote a fair learning model optimized for Equality of Opportunity as \textit{Fleoo}, while a model optimized for Statistical Parity is denoted as \textit{Flpar}. This distinction in terminology reflects the varying debiasing strategies employed to achieve different fairness criteria.

\subsection{Experiments}
To evaluate the effectiveness of various fairness intervention techniques, we conducted a series of experiments on the selected datasets. Each fairness method was applied to each GCN architectures, and the models were retrained using a range of fairness parameters. Specifically, for methods involving adjustable parameters, multiple models were trained with varying parameter values to explore the impact of different fairness constraints. For methods without adjustable parameters, such as edge weighting and embedding projection, a single model was trained for each dataset.

Following the retraining process, we assessed the performance of each model across multiple dimensions:
\begin{itemize}
\item \textbf{Privacy Concerns} We evaluated the vulnerability of the models to privacy attacks, privacy sensitive attribute inference (privacy leakage) and membership inference attacks. This involved determining the extent to which an adversary could infer the presence of a specific node in the training data based on the model's output.
\item \textbf{Fairness Metrics} We measured the fairness of the models using two key metrics: Equality of Opportunity and Statistical Parity. These metrics assess the disparity in model outcomes across different demographic groups defined by sensitive attributes. We also measured fairness leakage as accuracy of a fairness sensitive attribute inference attack.
\item \textbf{Accuracy} We evaluated the overall predictive accuracy of the models to understand the trade-offs between fairness and accuracy.
\end{itemize}
This comprehensive evaluation allowed us to analyze the impact of each fairness intervention technique on both privacy and fairness, while also considering the potential effects on the model's overall predictive performance.

%% file: 5_findings.tex
\section{Findings}
\subsection{Adversarial de-biasing}
\begin{figure}[!t]
    \centering
    \begin{subfigure}{0.48\textwidth}
        \includegraphics[width=\textwidth]{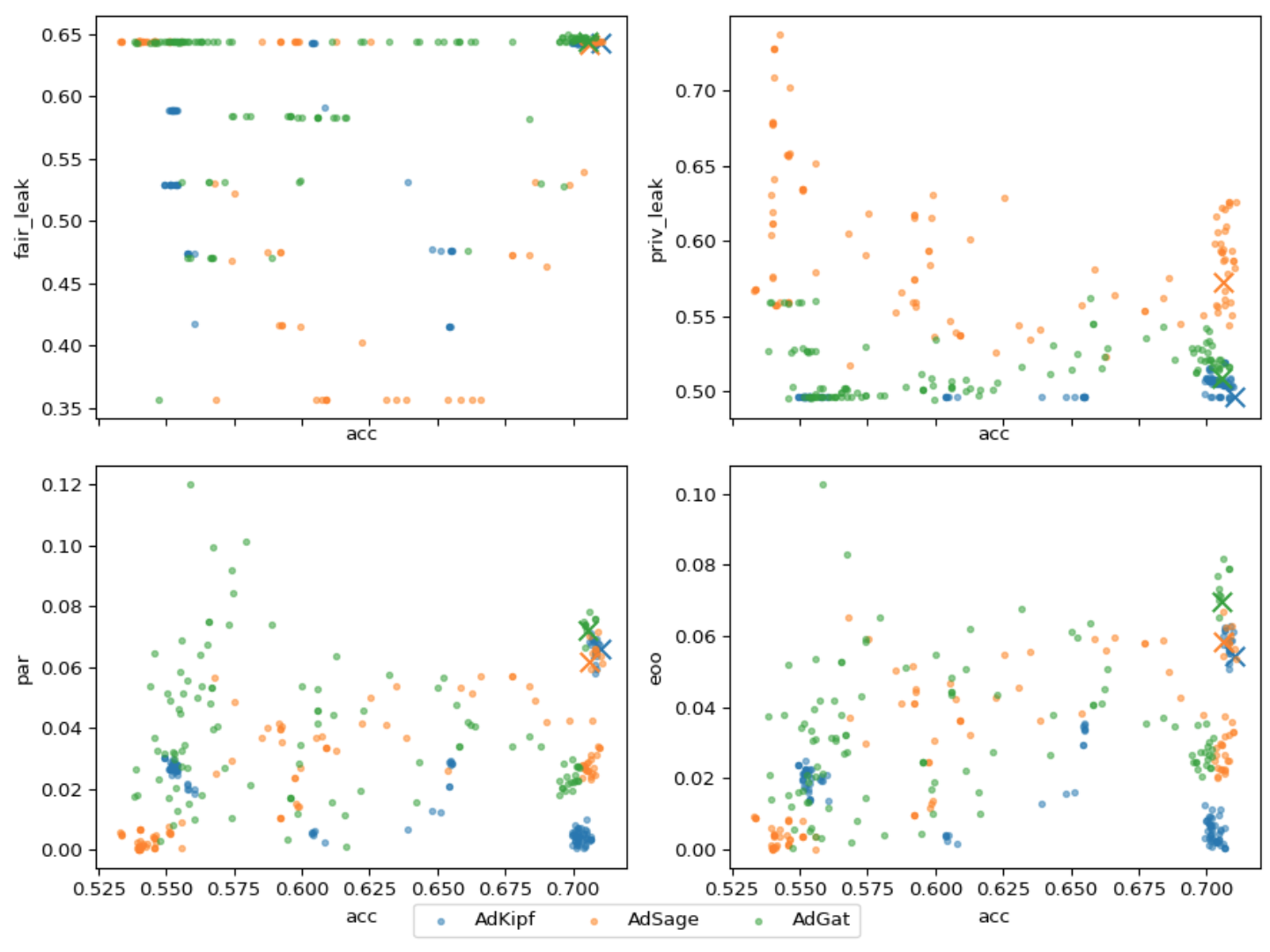}
        \caption{on Pokec z dataset}
    \end{subfigure}
    \hfill
    \begin{subfigure}{0.48\textwidth}
        \includegraphics[width=\textwidth]{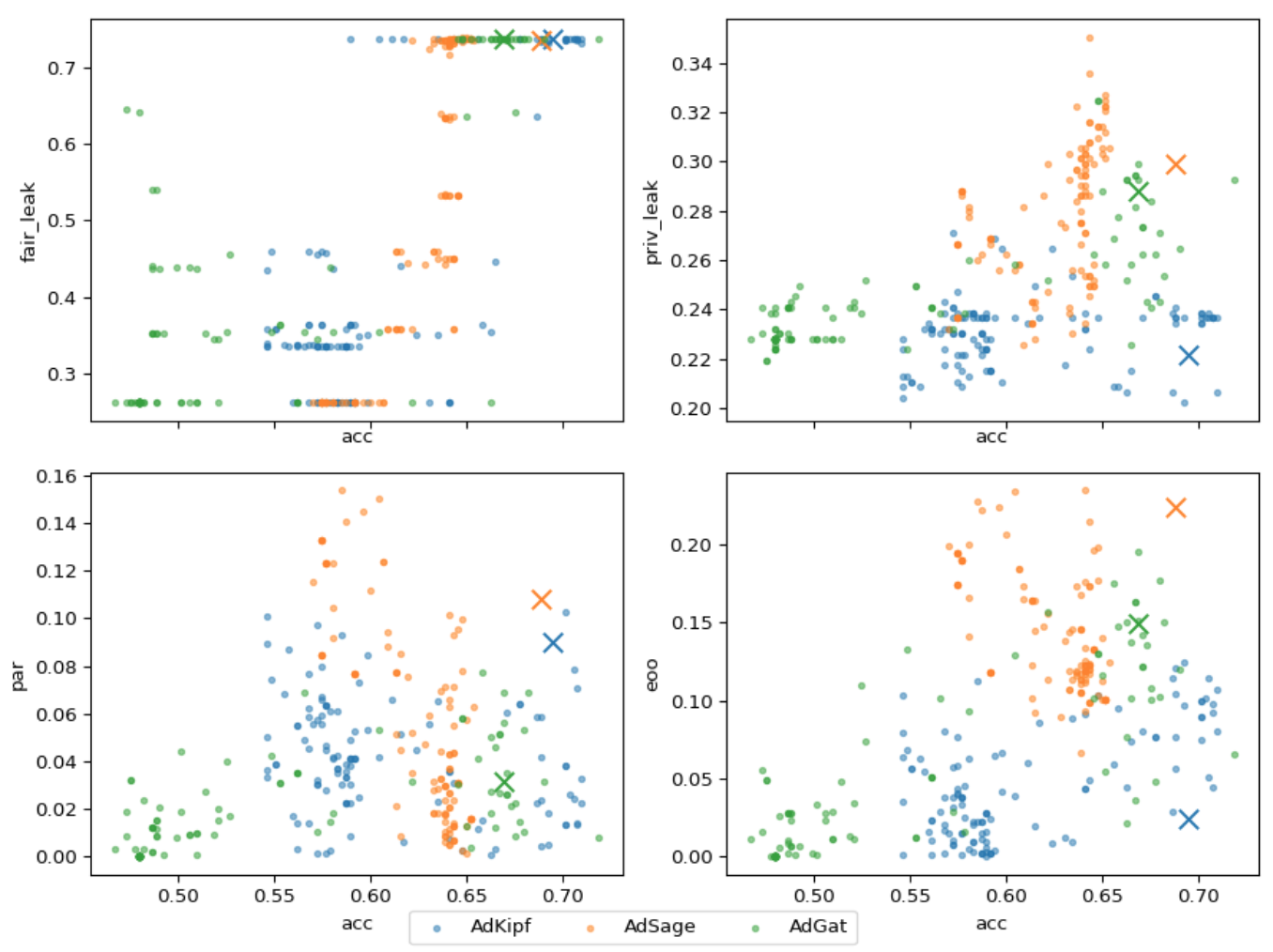}
        \caption{on NBA dataset}
    \end{subfigure}
    \caption{Performance of adversarial de-biasing with different $\alpha$ and $\beta$ parameters. X marks baseline model performance (without fairness countermeasures)}
    \label{fig:ad}
\end{figure}

Figure~\ref{fig:ad} illustrates the performance of various GNN architectures when combined with adversarial training for debiasing. Notably, the GCN architecture proposed by \citet{kipf2016semi} exhibits strong performance, achieving a low fairness score without compromising privacy or accuracy. This suggests that the GCN architecture is robust to adversarial debiasing techniques and can effectively mitigate bias while maintaining overall performance. In contrast, other architectures appear more sensitive to adversarial training, demonstrating a trade-off between fairness, privacy, and accuracy.  Furthermore, the consistent performance across all three datasets indicates that the observed trends are not dataset-specific but rather reflect inherent properties of the GNN architectures.

\begin{figure}[!t]
    \centering
    \includegraphics[width=0.7\textwidth]{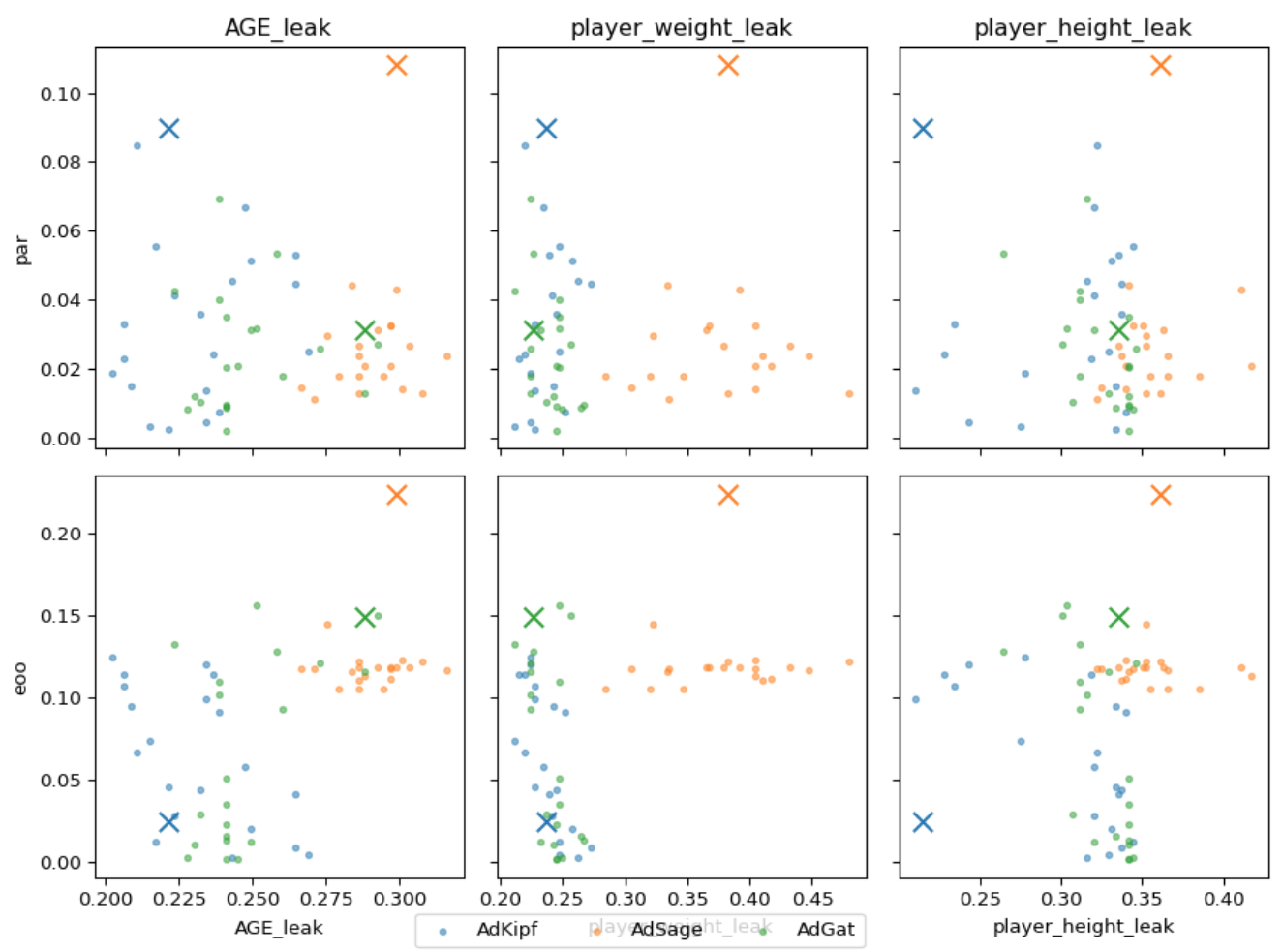}
    \caption{Different privacy sensitive attribues leakage under Adversarial debiasing}
    \label{fig:ad_nba_priv}
\end{figure}

To investigate the relationship between privacy leakage and fairness improvements, we analyzed the impact of varying fairness parameters ($\alpha$ and $\beta$) on the privacy vulnerability of different models. As depicted in Figure~\ref{fig:ad_nba_priv}, while the models exhibit varying degrees of privacy leakage, adversarial debiasing does not appear to have a significant overall impact on privacy. This suggests that the adversarial debiasing process, while effective in mitigating bias, does not generally increase or decrease the susceptibility of the models to privacy attacks. The observed variations in privacy leakage appear to be primarily attributed to the inherent characteristics of the different models rather than the influence of the adversarial debiasing technique.

\subsection{Other fairness countermeasures}
Given the observed heterogeneity in base model performance, we have opted to present the results stratified by both model architecture and dataset. This approach allows for a more nuanced analysis of the interaction between specific model characteristics and dataset properties. However, to facilitate comparison across different fairness-preserving measures, we present the results for each measure collectively, enabling a direct evaluation of their relative effectiveness within each model-dataset context.

\begin{figure}
    \centering
    \begin{subfigure}{0.48\textwidth}
        \centering
        \includegraphics[width=\textwidth]{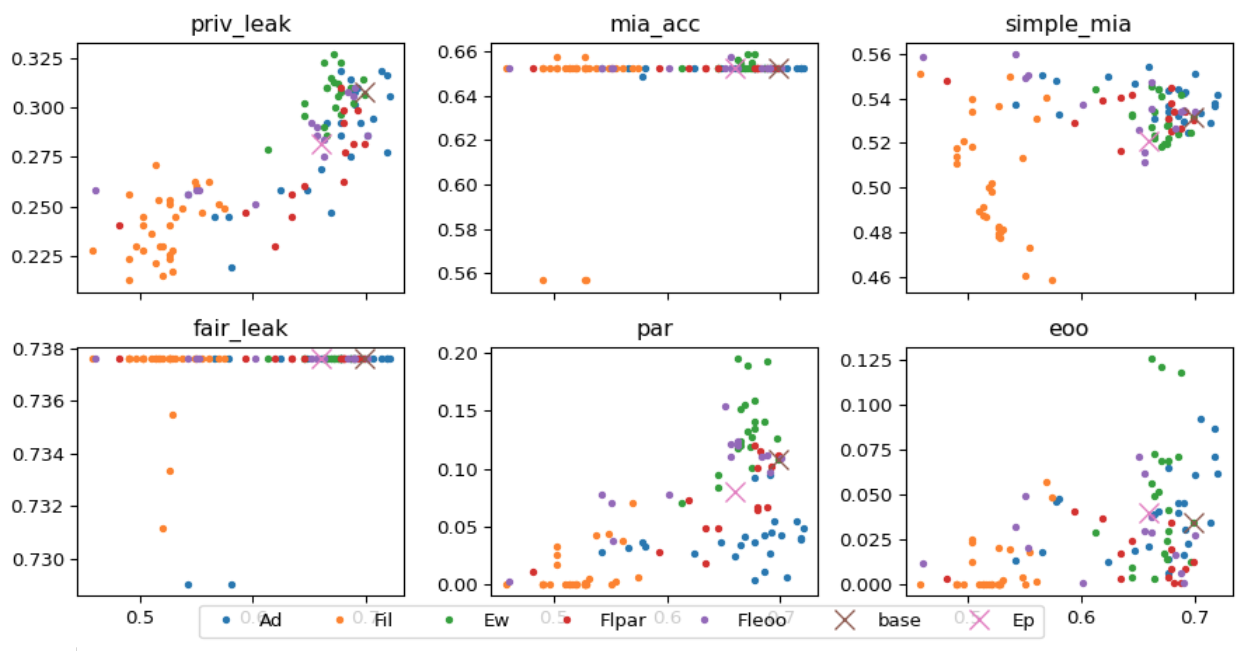}
        \caption{Kipf GCN}
    \end{subfigure}
    \hfil
    \begin{subfigure}{0.48\textwidth}
        \centering
        \includegraphics[width=\textwidth]{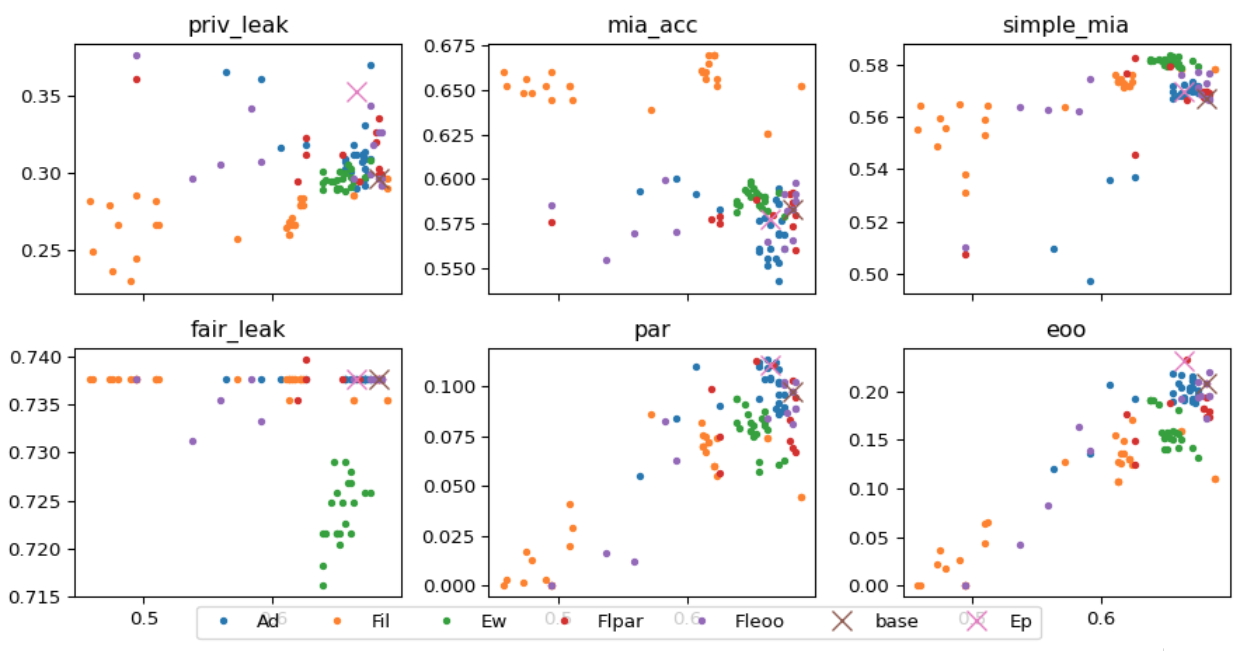}
        \caption{GraphSAGE}
    \end{subfigure}
    \hfil
    \begin{subfigure}{0.48\textwidth}
        \centering
        \includegraphics[width=\textwidth]{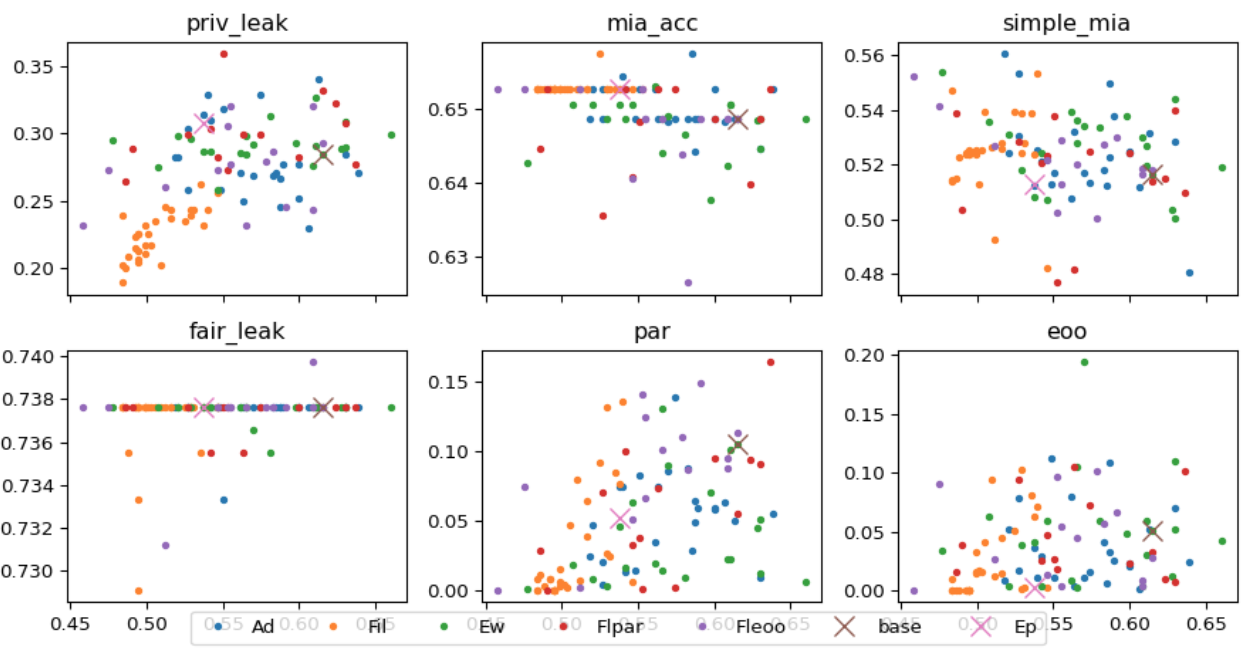}
        \caption{GIN}
    \end{subfigure}
    \hfil
    \begin{subfigure}{0.48\textwidth}
        \centering
        \includegraphics[width=\textwidth]{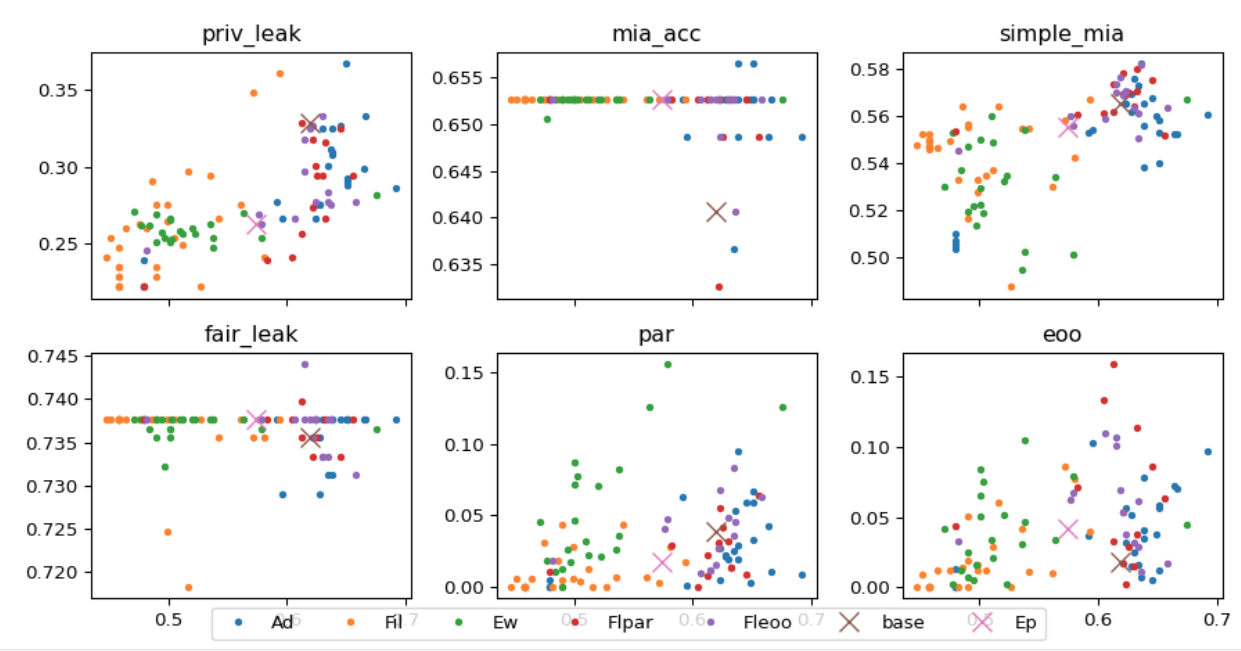}
        \caption{GAT}
    \end{subfigure}
    \caption{Performance of fairness countermeasures applied to different GCNs on nba dataset}
    \label{fig:nba_aggregate}
\end{figure}

\begin{figure}
    \centering
    \begin{subfigure}{0.48\textwidth}
        \centering
        \includegraphics[width=\textwidth]{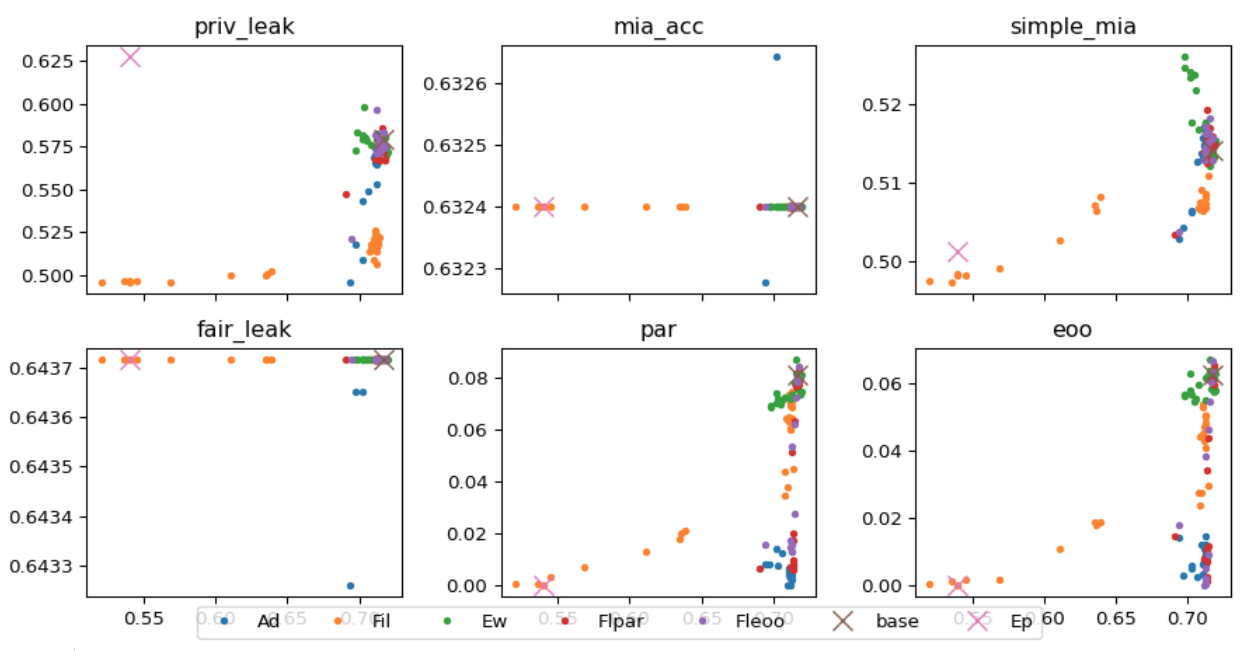}
        \caption{Kipf GCN}
    \end{subfigure}
    \hfil
    \begin{subfigure}{0.48\textwidth}
        \centering
        \includegraphics[width=\textwidth]{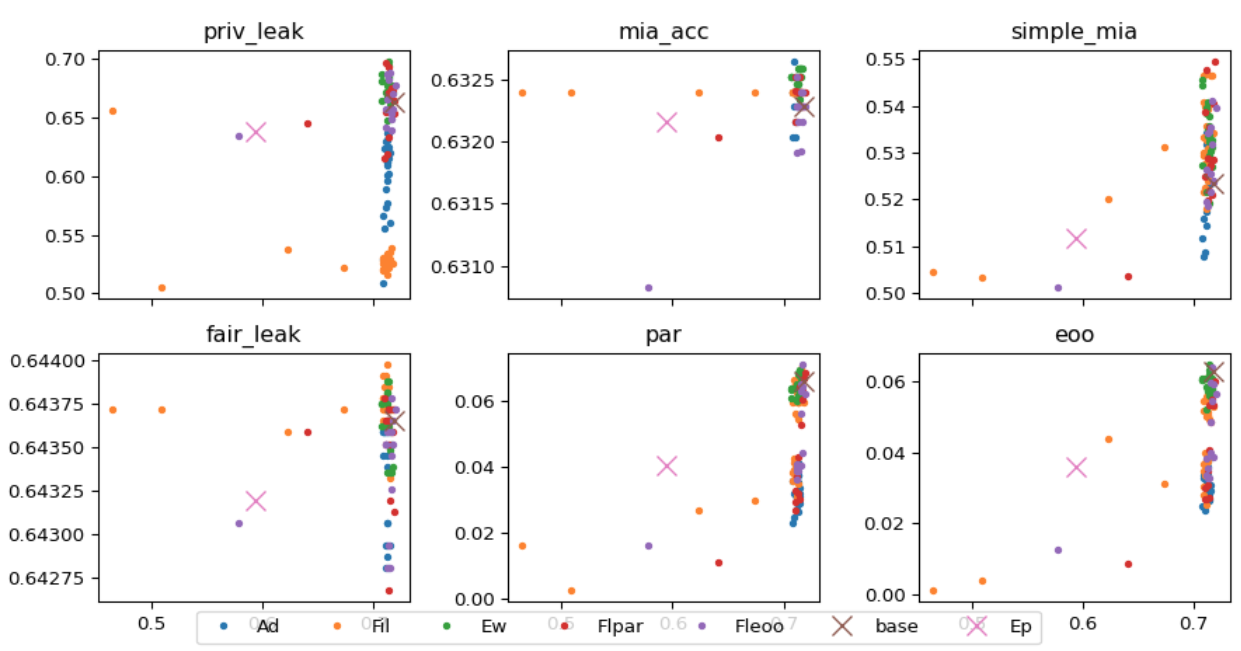}
        \caption{GraphSAGE}
    \end{subfigure}
    \hfil
    \begin{subfigure}{0.48\textwidth}
        \centering
        \includegraphics[width=\textwidth]{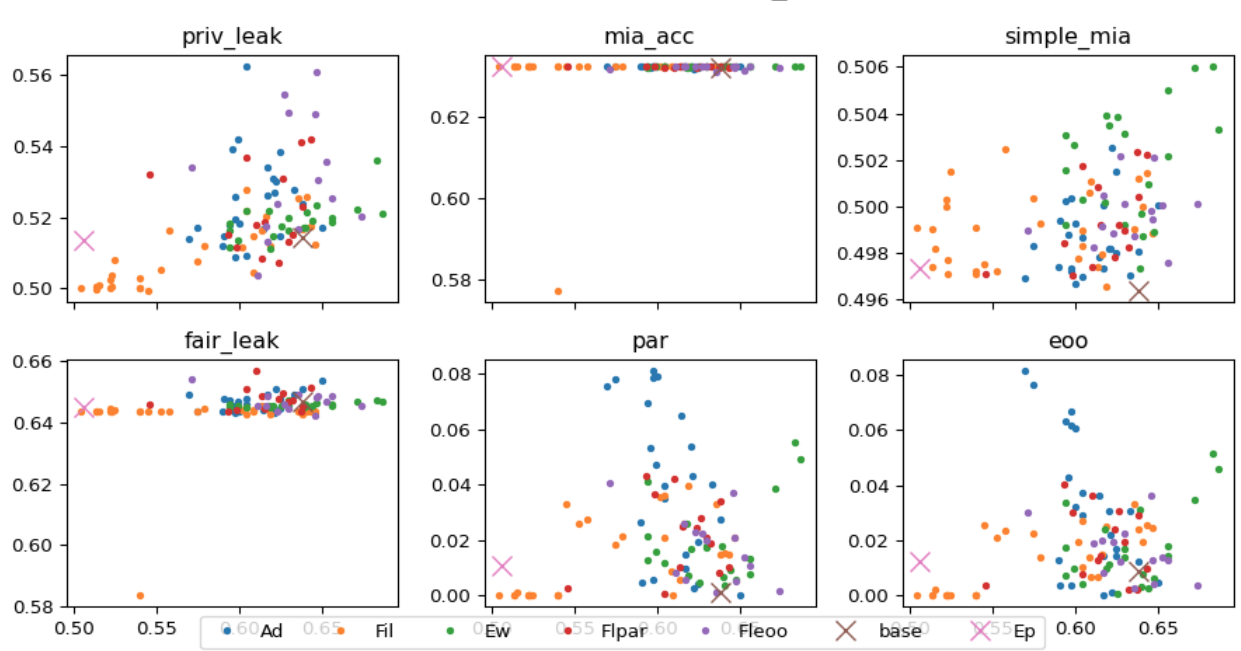}
        \caption{GIN}
    \end{subfigure}
    \hfil
    \begin{subfigure}{0.48\textwidth}
        \centering
        \includegraphics[width=\textwidth]{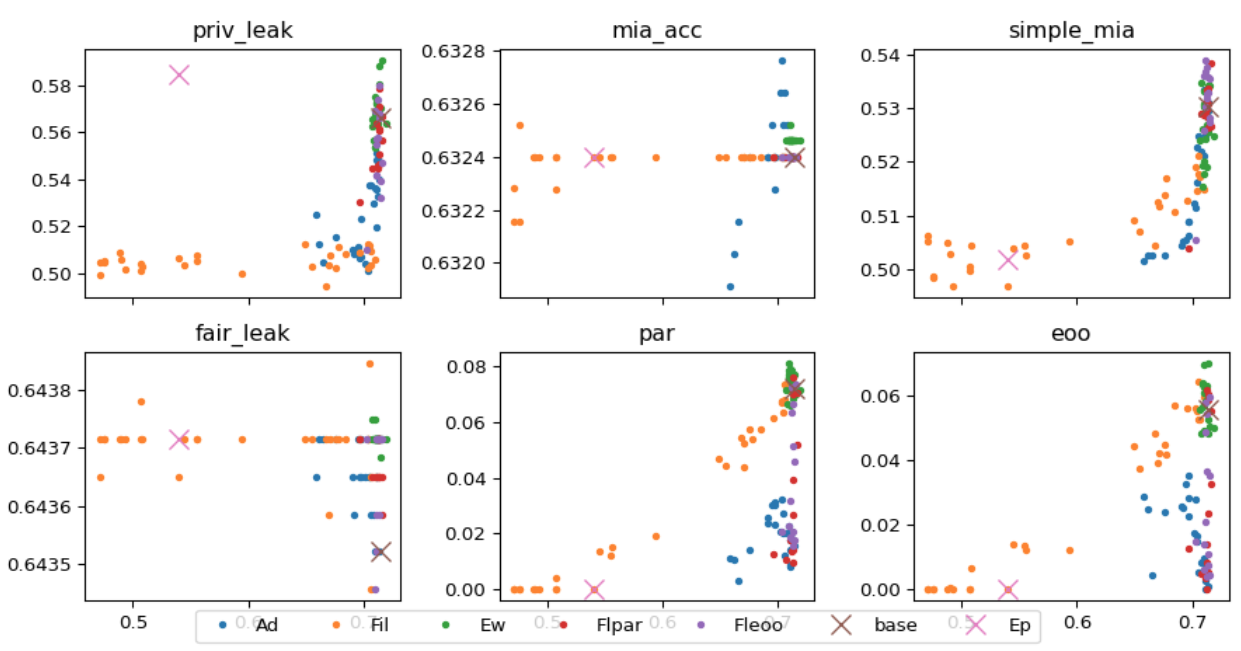}
        \caption{GAT}
    \end{subfigure}
    \caption{Performance of fairness countermeasures applied to different GCNs on Pokec z dataset}
    \label{fig:pokecz_aggregate}
\end{figure}

\begin{figure}
    \centering
    \begin{subfigure}{0.48\textwidth}
        \centering
        \includegraphics[width=\textwidth]{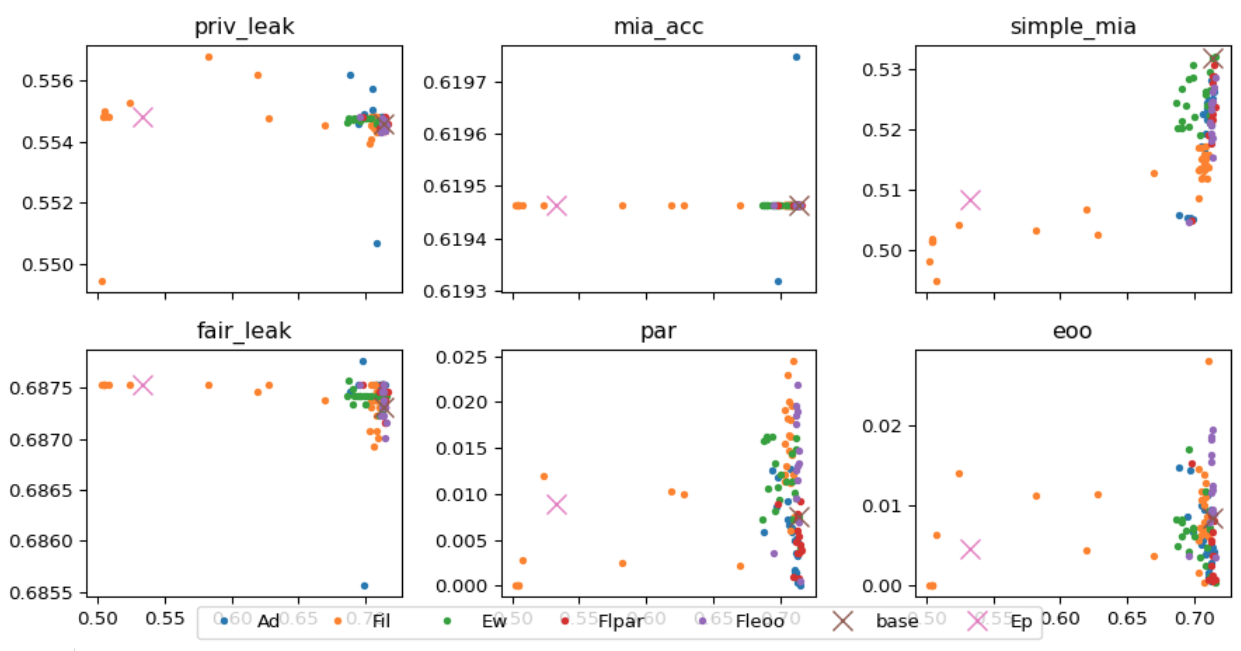}
        \caption{Kipf GCN}
    \end{subfigure}
    \hfil
    \begin{subfigure}{0.48\textwidth}
        \centering
        \includegraphics[width=\textwidth]{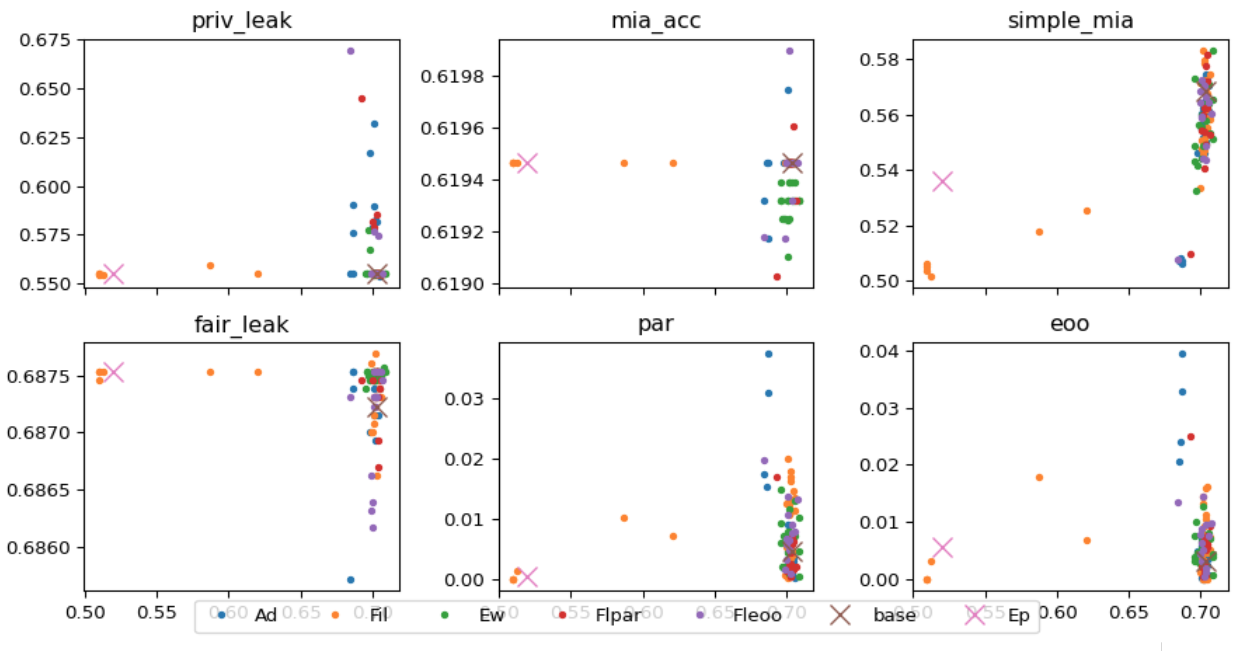}
        \caption{GraphSAGE}
    \end{subfigure}
    \hfil
    \begin{subfigure}{0.48\textwidth}
        \centering
        \includegraphics[width=\textwidth]{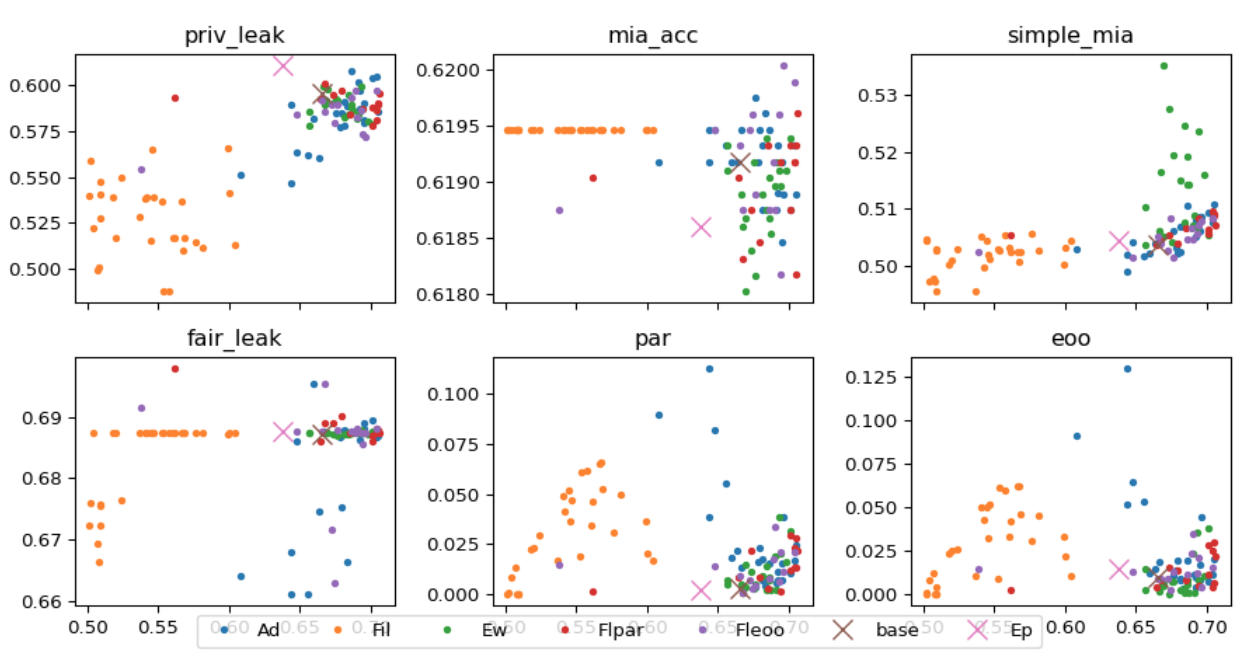}
        \caption{GIN}
    \end{subfigure}
    \hfil
    \begin{subfigure}{0.48\textwidth}
        \centering
        \includegraphics[width=\textwidth]{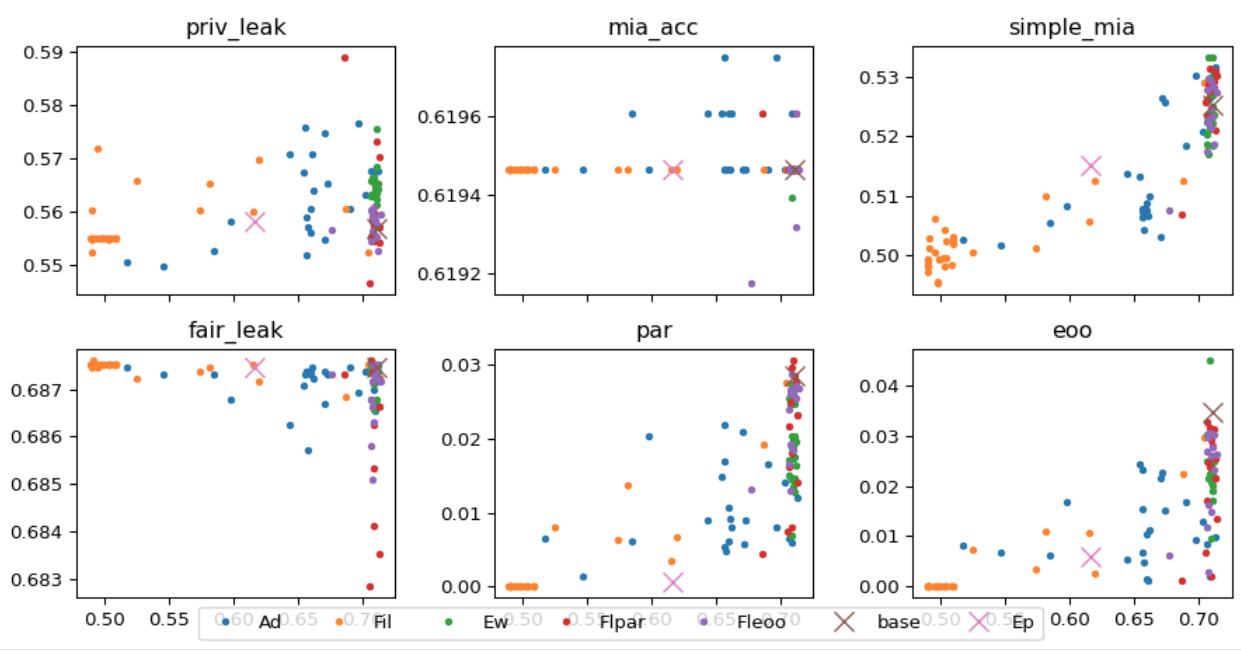}
        \caption{GAT}
    \end{subfigure}
    \caption{Performance of fairness countermeasures applied to different GCNs on Pokec n dataset}
    \label{fig:pokecn_aggregate}
\end{figure}

Figures \ref{fig:nba_aggregate}, \ref{fig:pokecz_aggregate}, and \ref{fig:pokecn_aggregate} provide a comparative analysis of the impact of various fairness-preserving measures on model performance across different datasets. Several key observations emerge from these figures:

\begin{itemize}
\item \textbf{Accuracy Trade-off:} Filtering and edge projection techniques, while effective in reducing privacy leakage, exhibit a significant decrease in accuracy. This highlights the inherent trade-off between privacy preservation and model performance.

\item \textbf{Consistency of Fairness and Privacy Metrics:} Fairness leakage and membership inference attacks (both simple and advanced) demonstrate consistent performance across all models, datasets, and fairness measurements. This suggests that these metrics are robust to variations in model architecture and data characteristics.

\item \textbf{Correlation between Privacy and Accuracy:} A strong correlation is observed between privacy leakage and accuracy, indicating that models with higher accuracy tend to exhibit greater vulnerability to privacy attacks. This underscores the challenge of simultaneously achieving high accuracy and strong privacy protection.

\item \textbf{Effectiveness of Fairness Models:} With the exception of filtering and edge projection, all fairness models achieve low equality of opportunity and statistical parity scores. Notably, adversarial debiasing, fair learning optimized for statistical parity, and edge weighting emerge as the most effective techniques for mitigating bias while maintaining reasonable accuracy.

\item \textbf{Model Robustness:} The Kipf GCN and GraphSAGE models exhibit the greatest robustness to fairness interventions, experiencing the lowest drop in accuracy when modified with various fairness methods. This suggests that these architectures are more amenable to debiasing techniques and can effectively incorporate fairness constraints without significant performance degradation.
\end{itemize}

\subsection{Combining Fairness Countermeasures}

This study investigates the potential for improved fairness outcomes by combining different fairness countermeasures. Specifically, we explored the integration of edge weighting with adversarial learning and fair learning, resulting in two novel techniques: Adversarial Edge Weighting (EwAd) and Fair Edge Weighting (EwFlpar).

Evaluation across the Pokec n, Pokec z, and NBA datasets revealed distinct performance patterns. While all models achieved near-zero scores for both equality of opportunity and statistical parity on the Pokec datasets, the NBA dataset exhibited more nuanced results.

\paragraph{Kipf GCN}

\begin{figure}
    \centering
    \begin{subfigure}{0.95\textwidth}
        \centering
        \includegraphics[width=\textwidth]{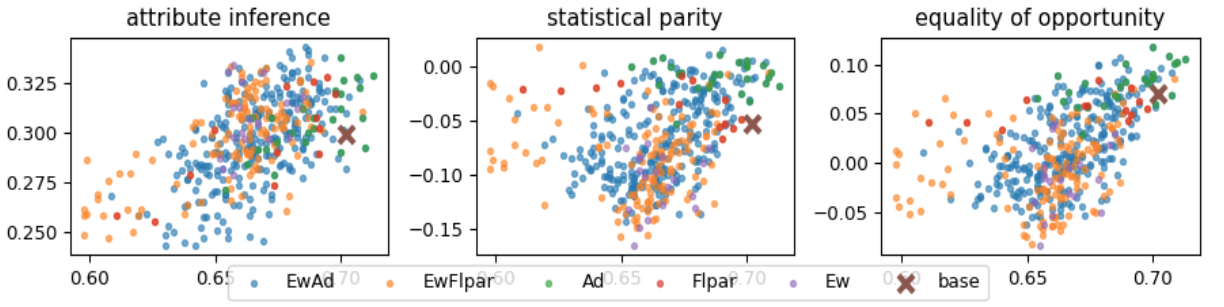}
        \caption{Kipf GCN}
        \label{fig:kipf_nba_combined}
    \end{subfigure}
    \begin{subfigure}{0.95\textwidth}
        \centering
        \includegraphics[width=\textwidth]{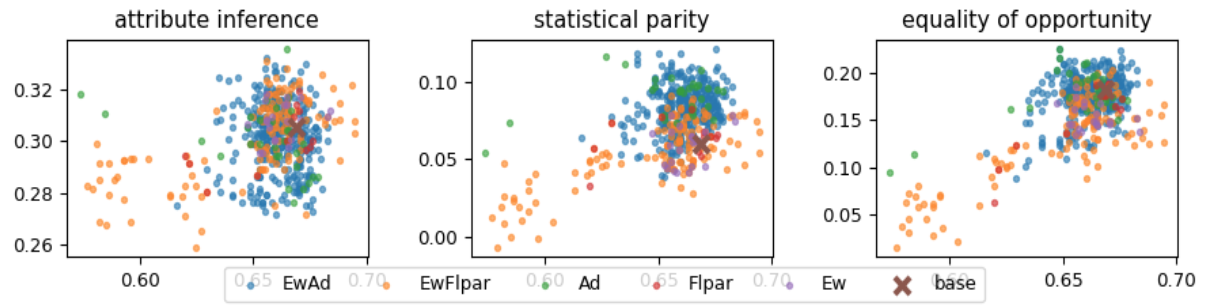}
        \caption{GraphSAGE}
        \label{fig:sage_nba_combined}
    \end{subfigure}
    \caption{Performance of combined fairness countermeasures on NBA dataset}
\end{figure}

Analysis of the Kipf GCN model, as illustrated in Figure~\ref{fig:kipf_nba_combined}, indicates that adversarial debiasing and adversarial edge weighting are most effective in minimizing equality of opportunity. In contrast, adversarial edge weighting and fair edge weighting excel in achieving low statistical parity scores, with adversarial debiasing alone demonstrating suboptimal performance in this regard.  Overall, adversarial edge weighting emerges as the most effective technique for the Kipf GCN, achieving optimal fairness outcomes while also minimizing privacy leakage.

\paragraph{GraphSAGE}

For the GraphSAGE model, Figure~\ref{fig:sage_nba_combined} demonstrates that fair edge weighting achieves superior performance, simultaneously maximizing accuracy and minimizing both statistical parity and equality of opportunity. This technique, along with adversarial edge weighting, also yields the lowest privacy leakage. Consequently, fair edge weighting is identified as the most effective fairness intervention for the GraphSAGE model.

These findings highlight the potential for combining different fairness countermeasures to achieve enhanced fairness outcomes. The optimal combination may vary depending on the specific model architecture and dataset characteristics, underscoring the importance of tailored debiasing strategies.

%% file: 6_conclusion.tex
\section{Conclusion}
In this paper, we conducted a comprehensive analysis of the impact of various fairness-preserving measures on graph neural networks, considering their effects on accuracy, privacy, and fairness metrics. Our experiments across diverse datasets and GNN architectures revealed key insights:
\begin{itemize}

\item \textbf{Trade-offs between fairness, privacy, and accuracy}: While several fairness interventions successfully mitigated bias, they often came at the cost of reduced accuracy or increased privacy leakage. This highlights the inherent challenges in achieving all three objectives simultaneously.

\item \textbf{Model robustness}: Certain GNN architectures, such as the Kipf GCN and GraphSAGE, exhibited greater robustness to fairness interventions, maintaining better accuracy and privacy preservation when debiasing techniques were applied. This suggests that the choice of GNN architecture plays a crucial role in achieving fair and private graph learning.

\item \textbf{Effectiveness of combined fairness measures}: Combining different fairness countermeasures, such as integrating edge weighting with adversarial learning or fair learning, demonstrated the potential for further enhancing fairness outcomes. This underscores the importance of exploring innovative combinations of existing techniques.

\item \textbf{Data-specific considerations}: The effectiveness of different fairness interventions varied across datasets, suggesting that the choice of debiasing techniques should be tailored to the specific characteristics of the data and the desired fairness objectives.

\end{itemize}

These findings contribute to a deeper understanding of the complex interplay between fairness, privacy, and accuracy in graph learning. Our study provides valuable insights for practitioners and researchers seeking to develop and deploy fair and private GNN models. Future research directions include exploring novel fairness interventions, developing more robust GNN architectures, and designing comprehensive evaluation frameworks that capture the multifaceted nature of fairness in graph-structured data.

\section{Future Work}

This study opens up several avenues for future research, including:

\begin{itemize}
\item \textbf{Joint Optimization of Fairness and Privacy:} A promising direction is to investigate the effectiveness of combining fairness and privacy-preserving measures. This would involve exploring how different fairness interventions interact with privacy-enhancing techniques and whether certain combinations yield synergistic benefits. Further research could examine whether the optimal combination of fairness and privacy measures varies across different fairness definitions, model architectures, or datasets.

\item \textbf{Automated Fairness and Privacy Optimization:} Developing a streamlined process for identifying the best combination of GNN architecture, fairness interventions, and privacy-preserving techniques for a specific use case is crucial. This could involve creating automated tools or frameworks that efficiently search the space of possible configurations and recommend optimal solutions based on the desired trade-offs between fairness, privacy, and accuracy.

\item \textbf{Dynamic and Adaptive Fairness:} Investigating fairness interventions that can adapt to evolving data distributions and dynamic graph structures is essential. This would involve developing methods that can continuously monitor and adjust fairness constraints to ensure ongoing fairness in real-world applications where data and relationships change over time.
\end{itemize}

By addressing these research questions, we can contribute to the development of more robust, fair, and privacy-preserving GNN models that can be reliably deployed in real-world scenarios.